\pdfoutput=1

\documentclass[11pt]{article}

\usepackage{emnlp2021}

\usepackage{times}
\usepackage{latexsym}

\usepackage[T1]{fontenc}

\usepackage[utf8]{inputenc}

\usepackage{booktabs}
\usepackage{multirow}
\usepackage{multicol}
\usepackage{graphicx}
\usepackage{tablefootnote}
\usepackage{threeparttable}
\usepackage{caption}
\usepackage{subcaption}
\usepackage{float}
\usepackage{siunitx}

\usepackage{microtype}

\title{FewshotQA: A simple framework for few-shot learning of question answering tasks 
using pre-trained text-to-text models}

\author{Rakesh Chada \\
  Amazon Alexa AI \\
  \texttt{rakchada@amazon.com} \\\And
  Pradeep Natarajan \\
  Amazon Alexa AI\\
  \texttt{natarap@amazon.com} \\}

\begin{document}
\maketitle
\begin{abstract}
    The task of learning from only a few examples (called a few-shot setting) is of key importance and relevance to a real-world setting. For question answering (QA), the current state-of-the-art pre-trained models typically need fine-tuning on tens of thousands of examples to obtain good results. Their performance degrades significantly in a few-shot setting (< 100 examples). To address this, we propose a simple fine-tuning framework that leverages pre-trained text-to-text models and is directly aligned with their pre-training framework. Specifically, we construct the input as a concatenation of the question, a mask token representing the answer span and a context. Given this input, the model is fine-tuned using the same objective as that of its pre-training objective. Through experimental studies on various few-shot configurations, we show that this formulation leads to significant gains on multiple QA benchmarks (an absolute gain of 34.2 F1 points on average when there are only 16 training examples). The gains extend further when used with larger models (Eg:- 72.3 F1 on SQuAD using BART-large with only 32 examples) and translate well to a multilingual setting . On the multilingual TydiQA benchmark, our model outperforms the XLM-Roberta-large by an absolute margin of upto 40 F1 points and an average of 33 F1 points in a few-shot setting (<= 64 training examples). We conduct detailed ablation studies to analyze factors contributing to these gains.
\end{abstract}

\section{Introduction}

\begin{figure*}[!ht]
\centering
\begin{subfigure}[b]{0.9\textwidth}
\centering
  \includegraphics[trim={0.7cm 0 0 0},scale=1.0,width=13.5cm]{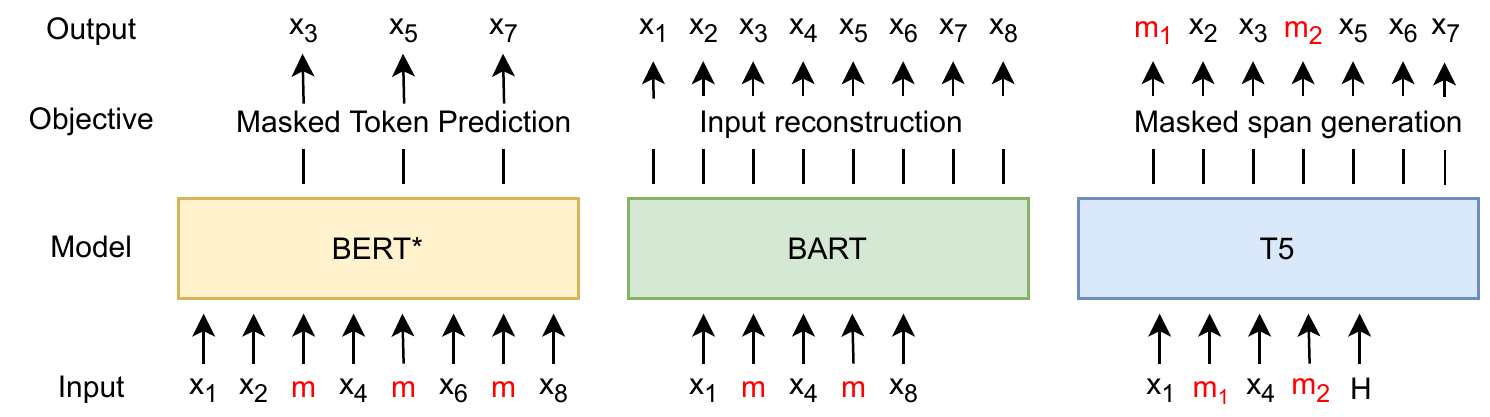}
  \caption{Comparison of pre-training frameworks. x$_i$ represent a single token at $i^{th}$ position. \textcolor{red}{m}  represents a special mask token. T5 uses a different mask token for each span masked (\textcolor{red}{m$_1$}, \textcolor{red}{m$_2$}).}
  \label{fig:pretrain}
\end{subfigure}

\definecolor{mypurple}{RGB}{97,65,195}

\begin{subfigure}[b]{0.9\textwidth}
\centering
  \includegraphics[trim={0.7cm 0 0 0},scale=0.94,width=13.5cm]{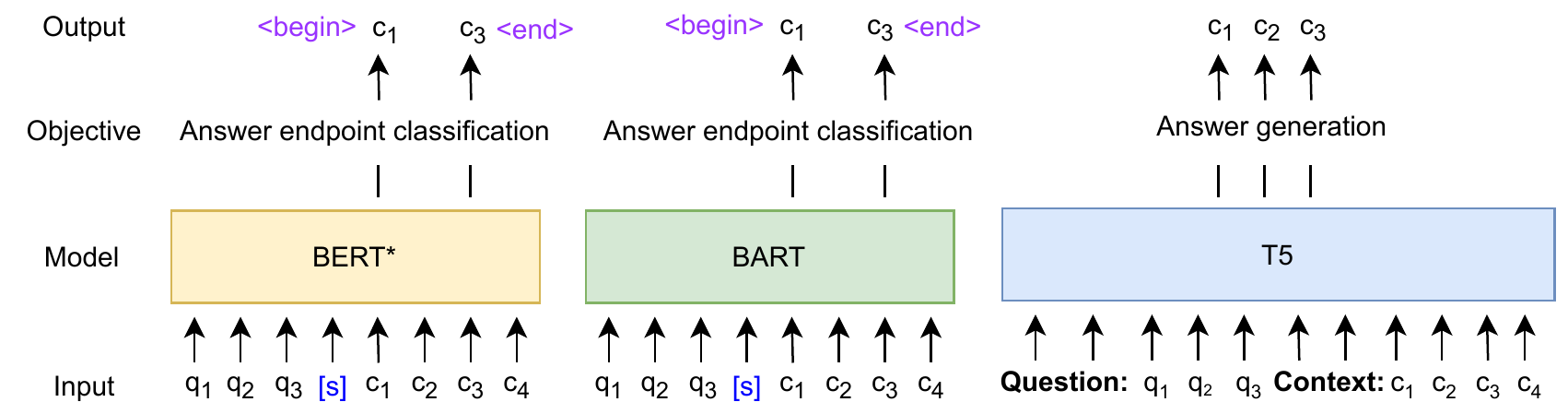}
  \caption{Comparison of existing fine-tuning frameworks. $q_i$ represent the question tokens. $c_i$ represents the context tokens. \textcolor{blue}{[s]} is a special delimiter symbol. \textcolor{mypurple}{<begin>}, \textcolor{mypurple}{<end>} represent the begin and end tokens for the answer span.}
  \label{fig:ftcur}
\end{subfigure}

\begin{subfigure}[b]{0.9\textwidth}
\centering
  \includegraphics[trim={0.7cm 0 0 0},scale=0.9,width=13.5cm]{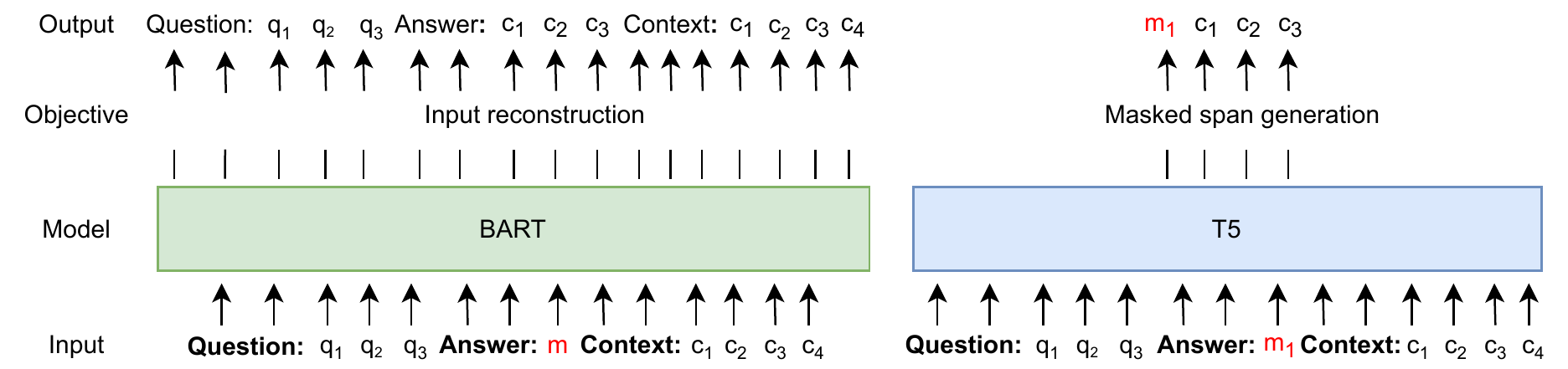}
  \caption{Our proposed FewshotQA fine-tuning framework. Note that the difference between this and the pre-training framework above are the inputs and outputs.}
  \label{fig:ftours}
\end{subfigure}

\caption{Comparison of different pre-training and fine-tuning systems for QA tasks with our proposed system. BERT$^*$ represents a class of BERT-like models that use the same pre-training objective.}
  \label{fig:comparison}
  
\end{figure*}

The task of question answering (QA) in Natural Language Processing typically involves producing an answer for a given question using a context that contains evidence to support the answer. The latest advances in pre-trained language models resulted in performance close to (and sometimes exceeding) a human performance when fine-tuned on several QA benchmarks \cite{devlin-etal-2019-bert}, \cite{NEURIPS2020_1457c0d6}, \cite{Bao2020UniLMv2PL}, \cite{JMLR:v21:20-074}. However, to achieve this result, these models need to be fine-tuned on tens of thousands of examples. In a more realistic and practical scenario, where only a handful of annotated training examples are available, their performance degrades significantly. For instance,~\cite{DBLP:conf/acl/RamKBGL20} show that, when only 16 training examples are available, the Roberta-base \cite{DBLP:journals/corr/abs-1907-11692} and SpanBERT-base \cite{joshi-etal-2020-spanbert} obtain a F1 score of 7.7 and 18.2 respectively on SQuAD \cite{rajpurkar-etal-2016-squad}. This is far lower than the F1 score of 90.3 and 92.0 when using the full training set of >100000 examples. Through experimental analysis, we observe that this degradation is majorly attributed to the disparities between fine-tuning and pre-training frameworks (a combination of the input-output design and the training objective). And to address this, we propose a fine-tuning framework (referred to as FewshotQA hereby) that is directly aligned with the pre-training framework, in terms of both the input-output design and the training objective. Specifically, we construct the input as a concatenation of the question, a mask token and context (in that order) and fine-tune a text-to-text pre-trained model using the same objective used during its pre-training to recover the answer. These text-to-text pre-trained model(s) were originally trained to recover missing spans of text in a given input sequence. And since our proposed fine-tuning setup is very much identical to the pre-training setup, this enables the model to make the best use of the pre-training "knowledge" for the fine-tuning task of question answering. 

The effectiveness of our FewshotQA system is shown in its strong results (an absolute average gain of 34.2 F1 points) on multiple QA benchmarks in a few-shot setting. We show that the gains extend further when used with larger sized models. We also test FewshotQA on a multilingual benchmark by replacing the pre-trained model with its multilingual counterpart and observe significant gains in comparison to a strong XLM-Roberta baseline (an absolute gain of 40 F1 points when there are only 16 training examples).

\section{Few-shot fine-tuning framework design}
Our proposed few-shot fine-tuning framework design involves a different choice of input-output design and the training objective than the current standard for QA fine-tuning frameworks. We provide a motivation for this design by comparison with the existing frameworks. Figure \ref{fig:comparison} illustrates this in detail. The pre-training framework is also pictured for comparison. Note that we focus on the bi-directional masked language models (MLMs) instead of the auto-regressive language models (such as GPT-2 \cite{Radford2019LanguageMA}) as the MLMs typically are deemed superior for QA tasks \cite{devlin-etal-2019-bert}, \cite{lewis-etal-2020-bart}.
\subsection{Pre-training} Figure \ref{fig:pretrain} illustrates the comparison between pre-training setups for three types of models. Firstly, there are BERT-style encoder-only models (referred to as BERT$*$) \cite{devlin-etal-2019-bert} that are pre-trained with the standard masked language modeling objective (also called a denoising objective) of predicting the masked tokens in an input sequence I. The masked tokens here typically correspond to a single word or a sub-word. Then, BART \cite{lewis-etal-2020-bart} uses a corrupted input reconstruction objective to recover the original input. The corruption involves replacing a span of multiple tokens with a mask token and sentence shuffling. Finally, T5 \cite{JMLR:v21:20-074} uses a masked span generation objective to predict the masked spans in an input. The input here is similar to that of BART where multiple spans are replaced with masked tokens. However, instead of generating the full input, only the masked spans are generated.
\subsection{Fine-tuning} Figure \ref{fig:ftcur} illustrates the fine-tuning setups for each of these models for the task of question answering. The input to both the BERT-style encoder-only models and BART is a concatenation of the question and the context. And both of them use a similar objective that encourages the model to predict the correct start and end positions of the answer in a given input. This is referred to as a span-selection objective. The input to T5 is also the concatenation of the question and the context. However, T5 uses an answer span generation objective to let the model directly generate the answer from scratch.
\subsection{Aligning the fine-tuning with pre-training}
The intuition behind aligning the fine-tuning and pre-training frameworks is that the model can make the best use of the "knowledge" obtained during pre-training phase in the fine-tuning phase. For question answering, the fine-tuning task involves predicting an answer span that could contain multiple tokens. This makes it non-trivial to align BERT$^*$ models for QA task during fine-tuning as their pre-training objectives let the model predict only a single word (or a sub-word) for a mask token. Similarly, it would require knowing the answer length in advance to have SpanBERT predict multi-masked tokens.\\
    Given that their pre-training objectives naturally involve multi-token span generation, BART and T5 make good candidates for this alignment. We further enhance the alignment by constructing the inputs (and outputs) to be similar to that of what the model sees during pre-training. This is done by appending a mask token (that should correspond to the answer in the target) as part of the input. This framework is illustrated in Figure \ref{fig:ftours}. We test the effectiveness of our formulation by putting it to test in various few-shot scenarios and observe significant gains.\\

Overall, we establish that the combination of text-to-text models and fine-tuning framework that is aligned with its pre-training counterpart makes a strong few-shot QA system. We now describe our experimental setup in Section \ref{modeling}.

\section{Modeling details} \label{modeling}
\subsection{Architecture} Our model follows the standard pre-trained text-to-text model architecture. It consists of a Transformer-based encoder and decoder models. We default to the "base" versions of the BART and T5 models as they contain a modest number of parameters (140M and 220M respectively) in comparison to the larger sized ones. For T5, we use the T5-V1.1 as the publicly released T5-V1.0 is fine-tuned on downstream tasks including question answering thereby contaminating it for our experimentation. BART-base consists of 6 encoder, 6 decoder layers with a hidden dimension of 768 and T5-base consists of 12 encoder, 12 decoder layers with a hidden dimension of 768 and a feed-forward hidden dimesion of 3072. We call the variants of BART, T5 used with our fine-tuning framework as FewshotBART and FewshotT5 respectively.

\begin{figure}[H]
\centering
  \includegraphics[trim={1cm 0 0 0},scale=0.65]{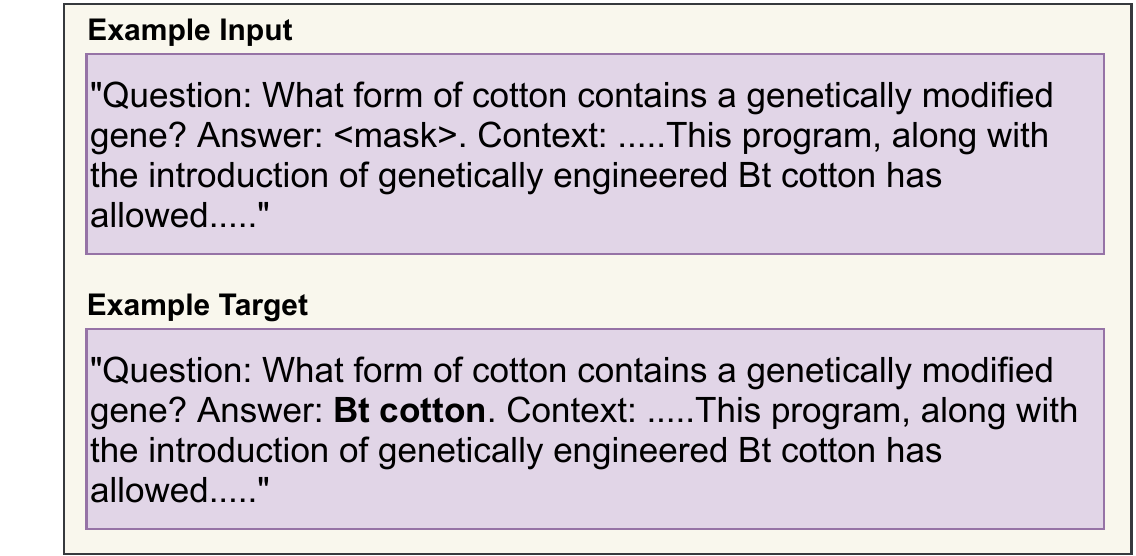}
  \caption{An example showing the input and target design for the FewshotQA model for BART.}
  \label{fig:example}
\end{figure}
\subsection{Input-output design \& fine-tuning objective} The input ($x_M$) to the model consists of a concatenation of three text sequences. The first ($x_q$) is a set of question tokens ($q$)  prefixed with the phrase ``Question:'', the second ($x_a$) is a mask token ($m$) prefixed with the phrase ``Answer:'' and the third ($x_c$) is a set of context tokens ($c$) prefixed with the phrase ``Context:''.\\\\
$x_q$ = \colorbox{gray!30}{Question: q}\\
$x_a$ = \colorbox{gray!30}{Answer: m}\\
$x_c$ = \colorbox{gray!30}{Context: c}\\
\[x_M = [x_q;x_a;x_c] \]

The target for the BART model ($y_{BART}$) is a concatenation of two text sequences, $x_q$ and $y_a$ where $y_a$ is the set of answer tokens prefixed with the phrase ``Answer:''\\\\
$y_a$ = \colorbox{gray!30}{Answer: a}\\
\[y_{BART} = [x_q;y_a] \]

The target for our T5 variant is a concatenation of a mask token $m$ and $y_a$ 
\[y_{T5} = [m;y_a] \]

An example of the input-target pairs from a dataset is shown in Figure \ref{fig:example}. 

The choice of text-to-text models in our system allows us to use to the standard encoder-decoder objective that maximizes the log likelihood of the text in the ground truth target from the output of the model. Formally, given the input $x_M$ and the target y (one of $y_{BART}$ and $y_{T5}$), the loss function L would now be:
\[ L(\theta) = \sum_{(x_M,y) \in (X_M,Y)}log(\prod_{i=1}^{n} P(y_i|y_{<i},x_M;\theta)) \]

Here, $X_M$ is the set of inputs, $Y$ is the set of targets, $n$ is the number of tokens in the target sequence. $y_i$ is the target token at timestep i. $y_{<i}$ represents all the target tokens preceding the timestep i. $P(y_i|y_{<i})$ represents the probability of the generating token $y_i$ given all the preceding ground truth tokens $y_{<i}$. And $\theta$ represents the parameters of the model.

We chose the order of concatenation in the input as question followed by a mask token followed by context as it enables us to run the generation process for a far fewer number of steps before an answer is generated. (see \textbf{Generation Strategy} section below). The context can be quite long so generating the entire context auto-regressively before generating an answer would be inefficient and cause a degradation in performance.

\begin{table*}[t]
\small
\centering
\begin{tabular}{@{}lcccccccc@{}}
\toprule
\textbf{Model} & \small\textbf{SQuAD} & \small\textbf{TriviaQA} & \small\textbf{NQ} & \small\textbf{NewsQA} & \small\textbf{SearchQA} & \small\textbf{HotpotQA} & \small\textbf{BioASQ} & \small\textbf{TbQA} \\
\midrule
\small\textit{16 Examples}  & & & & & & & &  \\
\midrule
BART & 10.44\textpm5.9 & 2.5\textpm2.1 & 13.3\textpm2.4 & 2.9\textpm1.4 & 5.3\textpm2.4 & 5.6\textpm2.4 & 10.3\textpm2.9 & 1.7\textpm1.2 \\
FewshotBART & \textbf{55.5}\textpm2.0 & \textbf{50.5}\textpm1.0 & \textbf{46.7}\textpm2.3 & \textbf{18.9}\textpm1.8 & \textbf{39.8}\textpm0.04 & \textbf{45.1}\textpm1.5 & \textbf{49.4}\textpm0.02 & \textbf{19.9}\textpm1.25 \\
\midrule
\small\textit{128 Examples}  & & & & & & & &  \\
\midrule
BART & 42.2\textpm3.4 & 11.1\textpm0.9 & 24.6\textpm1.9 & 18.1\textpm1.3 & 17.8\textpm2.3 & 23.0\textpm2.5 & 39.1\textpm2.8 & 9.2\textpm0.7 \\
FewshotBART & \textbf{68.0}\textpm0.3 & \textbf{50.1}\textpm1.8 & \textbf{53.9}\textpm0.9 & \textbf{47.9}\textpm1.2 & \textbf{58.1}\textpm1.4 & \textbf{54.8}\textpm0.8 & \textbf{68.5}\textpm1.0 & \textbf{29.7}\textpm2.4 \\
\bottomrule
\end{tabular}
\caption{Comparison of F1 scores across all datasets for the standard QA fine-tuning objective (BART) vs the proposed aligned fine-tuning objective (FewshotBART). The value after \textpm{ } indicates the standard deviation across 5 runs with different seeds. NQ stands for Natural Questions. TbQA stands for TextbookQA.}
\label{tab:alignment_results}
\end{table*}

\begin{table*}[t]
\small
\centering
\begin{tabular}{@{}lcccccccc@{}}
\toprule
\textbf{Model} & \small\textbf{SQuAD} & \small\textbf{TriviaQA} & \small\textbf{NQ} & \small\textbf{NewsQA} & \small\textbf{SearchQA} & \small\textbf{HotpotQA} & \small\textbf{BioASQ} & \small\textbf{TbQA} \\
\midrule
\small\textit{16 Examples}  & & & & & & & &  \\
\midrule
FewshotBART & \textbf{55.5}\textpm2.0 & 50.5\textpm1.0 & \textbf{46.7}\textpm2.3 & \textbf{38.9}\textpm0.7 & \textbf{39.8}\textpm0.04 & \textbf{45.1}\textpm1.5 & \textbf{49.4}\textpm0.02 & 19.9\textpm1.25 \\
FewshotBARTL & \textbf{\textcolor{blue}{68.9}}\textpm2.7 & \textbf{\textcolor{blue}{65.2}}\textpm1.8 & \textbf{\textcolor{blue}{60.4}}\textpm2.0 &
\textbf{\textcolor{blue}{48.4}}\textpm2.2 & \textbf{\textcolor{blue}{47.8}}\textpm5.4 & \textbf{\textcolor{blue}{58.0}}\textpm1.8 & \textbf{\textcolor{blue}{63.0}}\textpm1.1 & \textbf{\textcolor{blue}{37.7}}\textpm3.7 \\
FewshotT5 & 47.8\textpm6.9 & \textbf{50.6}\textpm4.9 & 28.5\textpm14.5 & 26.8\textpm2.7 & 37.0\textpm3.3 & 44.9\textpm3.5 & 46.3\textpm5.9 & \textbf{25.9}\textpm5.0 \\
RoBERTa & ~~7.7\textpm4.3 & ~~7.5\textpm4.4 & 17.3\textpm3.3 & ~~1.4\textpm0.8 & ~~6.9\textpm2.7 & 10.5\textpm2.5 & 16.7\textpm7.1 & ~~3.3\textpm2.1 \\
SpanBERT & 18.2\textpm6.7 & 11.6\textpm2.1 &19.6\textpm3.0 & ~~7.6\textpm4.1 & 13.3\textpm6.0 & 12.5\textpm5.5 & 15.9\textpm4.4 & ~~7.5\textpm2.9 \\
Splinter & 54.6\textpm6.4 & 18.9\textpm4.1 & 27.4\textpm4.6 & 20.8\textpm2.7 & 26.3\textpm3.9 & 24.0\textpm5.0 & 28.2\textpm4.9 & 19.4\textpm4.6 \\
\midrule
\small\textit{32 Examples}  & & & & & & & &  \\
\midrule
FewshotBART & 56.8\textpm2.1 & \textbf{52.5}\textpm0.7 & \textbf{50.1}\textpm1.1 & \textbf{40.4}\textpm1.5 & 41.8\textpm0.02 & 47.9\textpm1.4 & 52.3\textpm0.02 & 22.7\textpm2.3 \\
FewshotBARTL & \textbf{\textcolor{blue}{72.3}}\textpm1.0 & \textbf{\textcolor{blue}{65.1}}\textpm1.2 & \textbf{\textcolor{blue}{61.5}}\textpm1.7 & \textbf{\textcolor{blue}{51.7}}\textpm1.7 & \textbf{\textcolor{blue}{58.3}}\textpm1.5 & \textbf{\textcolor{blue}{60.4}}\textpm0.2 & \textbf{\textcolor{blue}{67.8}}\textpm1.0 & \textbf{\textcolor{blue}{37.7}}\textpm9.8 \\
FewshotT5 & 56.6\textpm1.5 & 50.2\textpm9.0 & 37.5\textpm12.5 & 33.2\textpm4.6 & \textbf{48.4}\textpm5.6 & \textbf{53.6}\textpm1.4 & \textbf{57.7}\textpm4.2 & \textbf{29.8}\textpm2.6 \\
RoBERTa &18.2\textpm5.1 &10.5\textpm1.8 &22.9\textpm0.7 &~~3.2\textpm1.7 &13.5\textpm1.8 &10.4\textpm1.9 &23.3\textpm6.6 &4.3\textpm0.9 \\
SpanBERT &25.8\textpm7.7 &15.1\textpm6.4 &25.1\textpm1.6 &~~7.2\textpm4.6 &14.6\textpm8.5 &13.2\textpm3.5 &25.1\textpm3.3 &~~7.6\textpm2.3 \\
Splinter & \textbf{59.2}\textpm2.1 & 28.9\textpm3.1 & 33.6\textpm2.4 & 27.5\textpm3.2 & 34.8\textpm1.8 & 34.7\textpm3.9 & 36.5\textpm3.2 & 27.6\textpm4.3 \\
\midrule
\small\textit{64 Examples}  & & & & & & & &  \\
\midrule
FewshotBART & 61.5\textpm2.3 & 50.8\textpm2.2 & \textbf{53.0}\textpm0.5 & \textbf{42.7}\textpm2.2 & 46.1\textpm2.9 & 51.2\textpm1.0 & 61.8\textpm2.8 & 27.6\textpm1.8 \\
FewshotBARTL & \textbf{\textcolor{blue}{73.6}}\textpm1.9 & \textbf{\textcolor{blue}{64.6}}\textpm1.4 & \textbf{\textcolor{blue}{63.0}}\textpm2.1 & \textbf{\textcolor{blue}{53.5}}\textpm0.9 & \textbf{\textcolor{blue}{65.5}}\textpm2.4 & \textbf{\textcolor{blue}{62.9}}\textpm1.6 & \textbf{\textcolor{blue}{73.9}}\textpm0.8 & \textbf{\textcolor{blue}{45.0}}\textpm1.7 \\
FewshotT5 & 57.2\textpm5.6 & \textbf{52.4}\textpm5.9 & 48.6\textpm2.1 & 40.2\textpm4.1 & \textbf{54.4}\textpm3.0 & \textbf{56.3}\textpm2.9 & \textbf{63.8}\textpm2.5 & 32.1\textpm2.7 \\
RoBERTa &28.4\textpm1.7 &12.5\textpm1.4 &24.2\textpm1.0 &~~4.6\textpm2.8 &19.8\textpm2.4 &15.0\textpm3.9 &34.0\textpm1.8 &~~5.4\textpm1.1 \\
SpanBERT &45.8\textpm3.3 &15.9\textpm6.4 &29.7\textpm1.5 &12.5\textpm4.3 &18.0\textpm4.6 &23.3\textpm1.1 &35.3\textpm3.1 &13.0\textpm6.9 \\
Splinter & \textbf{65.2}\textpm1.4 & 35.5\textpm3.7 & 38.2\textpm2.3 & 37.4\textpm1.2 & 39.8\textpm3.6 & 45.4\textpm2.3 & 49.5\textpm3.6 & \textbf{35.9}\textpm3.1 \\
\midrule
\small\textit{128 Examples}  & & & & & & & &  \\
\midrule
FewshotBART & 68.0\textpm0.3 & 50.1\textpm1.8 & \textbf{53.9}\textpm0.9 & \textbf{47.9}\textpm1.2 & \textbf{58.1}\textpm1.4 & 54.8\textpm0.8 & \textbf{68.5}\textpm1.0 & 29.7\textpm2.4 \\
FewshotBARTL & \textbf{\textcolor{blue}{79.4}}\textpm1.5 & \textbf{\textcolor{blue}{65.8}}\textpm0.9 & \textbf{\textcolor{blue}{64.3}}\textpm1.3 & \textbf{\textcolor{blue}{57.0}}\textpm0.9 & \textbf{\textcolor{blue}{67.7}}\textpm1.0 & \textbf{\textcolor{blue}{75.1}}\textpm1.5 & \textbf{\textcolor{blue}{75.0}}\textpm1.5 & \textbf{\textcolor{blue}{48.4}}\textpm2.7 \\
FewshotT5 & 64.6\textpm6.1 & \textbf{51.7}\textpm3.1 & 47.0\textpm4.6 & 40.0\textpm1.9 & 57.0\textpm4.5 & \textbf{56.1}\textpm3.7 & 68.2\textpm3.6 & 33.6\textpm2.1 \\
RoBERTa &43.0\textpm7.1 &19.1\textpm2.9 &30.1\textpm1.9 &16.7\textpm3.8 &27.8\textpm2.5 &27.3\textpm3.9 &46.1\textpm1.4 &~~8.2\textpm1.1 \\
SpanBERT &55.8\textpm3.7 &26.3\textpm2.1 &36.0\textpm1.9 &29.5\textpm7.3 &26.3\textpm4.3 &36.6\textpm3.4 &52.2\textpm3.2 &20.9\textpm5.1 \\
Splinter & \textbf{72.7}\textpm1.0 & 44.7\textpm3.9 & 46.3\textpm0.8 & 43.5\textpm1.3 & 47.2\textpm3.5 & 54.7\textpm1.4 & 63.2\textpm4.1 & \textbf{42.6}\textpm2.5 \\
\bottomrule
\end{tabular}
\caption{F1 scores across all datasets and training set sizes. Bold indicates the best result. FewshotBARTL's results are colored blue as it has many more parameters (406M) than Splinter (110M). FewshotBART has a comparable number of parameters (130M). FewshotT5 has 220M parameters.}
\label{tab:fewshot_results}
\end{table*}

\subsection{Generation strategy} During both validation and testing, the model is provided the special start token as input and asked to generate tokens in an auto-regressive manner for a fixed number of steps. For BART, since the question and answer tokens are at the beginning of the sequence in the input and the model is trained to reconstruct the input, we just need to generate until the answer is generated. In practice, for the datasets experimented, the generation length of 50 is sufficient to generate the answer. We stop the generation once these 50 tokens are generated. For T5, the generation length is set to 25 as only the answer is generated. For both, we use greedy decoding with a beam size = 1.
\subsection{Answer extraction} Once the outputs are generated, the answers are then extracted via a simple post-processing rule that extracts the answer part of the generation. The use of a fixed input pattern (Question: a Answer: a Context: c) makes this step a simple deterministic one without the need for additional heuristics.
\subsection{Multilingual extension}
Our fine-tuning framework can be extended to a multilingual question answering setting by switching the pretrained model with its multilingual counterpart. We experiment by replacing BART-base model with mBART-50 model that was pre-trained with the same objective as that of BART on a multilingual corpus. The rest of the components, fine-tuning objective and the answer extraction, remain the same as that of the FewshotBART. We call this model FewshotmBART.

\subsection{Hyperparameters}
We use Adam optimizer with a learning rate of 2e-5. We use a training batch size of 4. We don't use learning rate scheduling. For evaluation on the test set, we pick the best model based off the development set performance. The maximum sequence length is set to the 99th percentile length of all sequence lengths in the development set. We train for a total of 35 epochs or 1000 steps (whichever is the maximum).

\section{Experiments}

\subsection{Datasets}

We follow~\cite{DBLP:conf/acl/RamKBGL20} and choose the datasets sampled from the MRQA shared task \cite{fisch-etal-2019-mrqa} for our few-shot experiments. We also use the same train and test splits provided in ~\cite{DBLP:conf/acl/RamKBGL20} for fine-tuning and evaluating our models. However, instead of fine-tuning for a fixed number of iterations, we use a development set to determine the best checkpoint to use for testing. These datasets contain 5000 to 17000 test examples.\\\\
\textbf{Development data split}: To cater to a realistic and a practical setting for a few-shot scenario, we pick the development set to be the same size as of the training set. As reported in~\cite{gao-etal-2021-making}, having access to the full development set during training would create an unrealistic few-shot setting. We also make sure there is no overlap between training and development sets.

Below, we describe results for several experiments conducted on MRQA few-shot datasets. We run each experiment five times using five different random seeds. And we report the mean and standard deviation of the results for each run.

\begin{figure*}[!ht]
     \centering
     \begin{subfigure}[b]{0.49\textwidth}
         \centering
         \includegraphics[width=\textwidth]{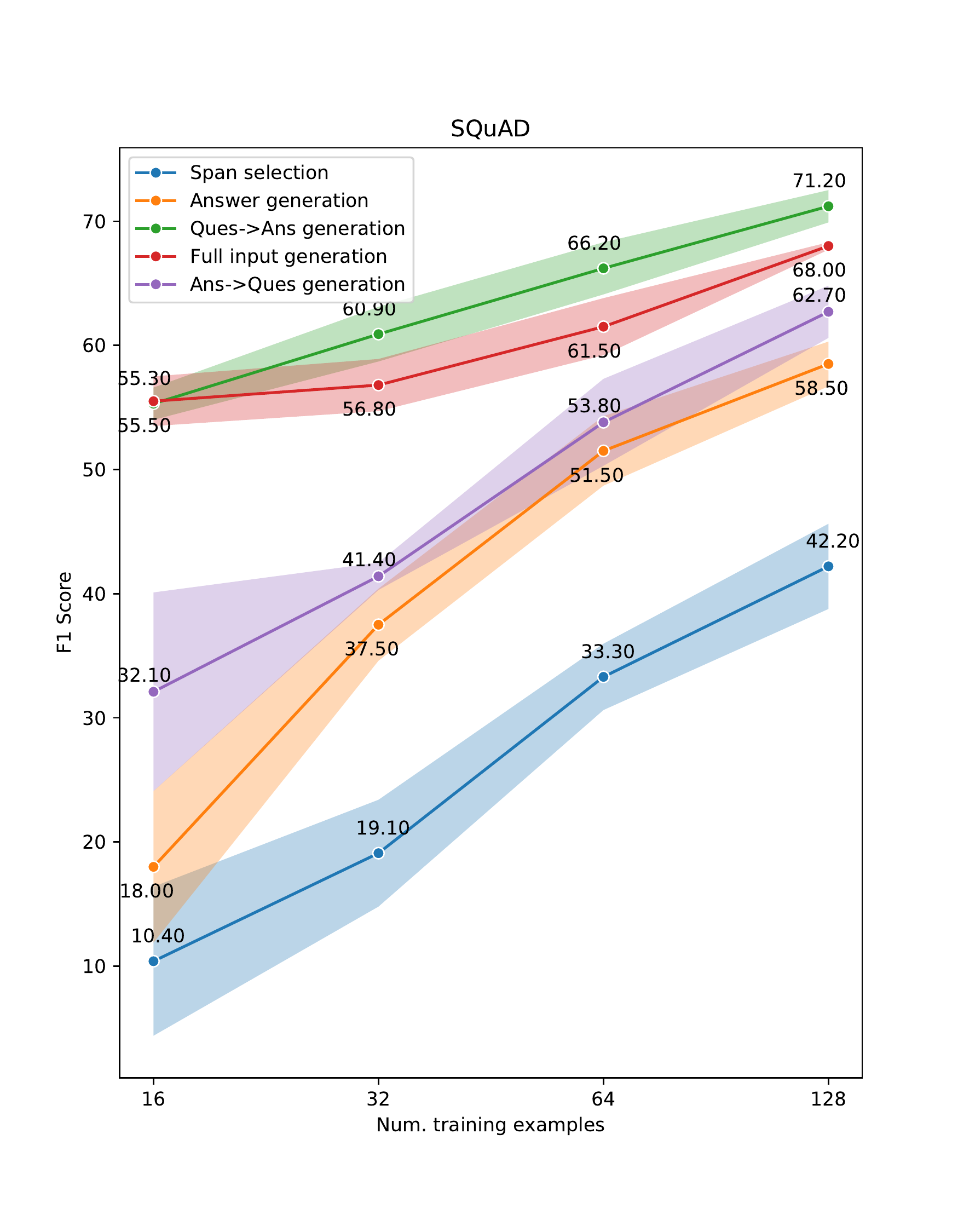}
         \label{fig:tydiqa}
     \end{subfigure}
     \begin{subfigure}[b]{0.49\textwidth}
         \centering
         \includegraphics[width=\textwidth]{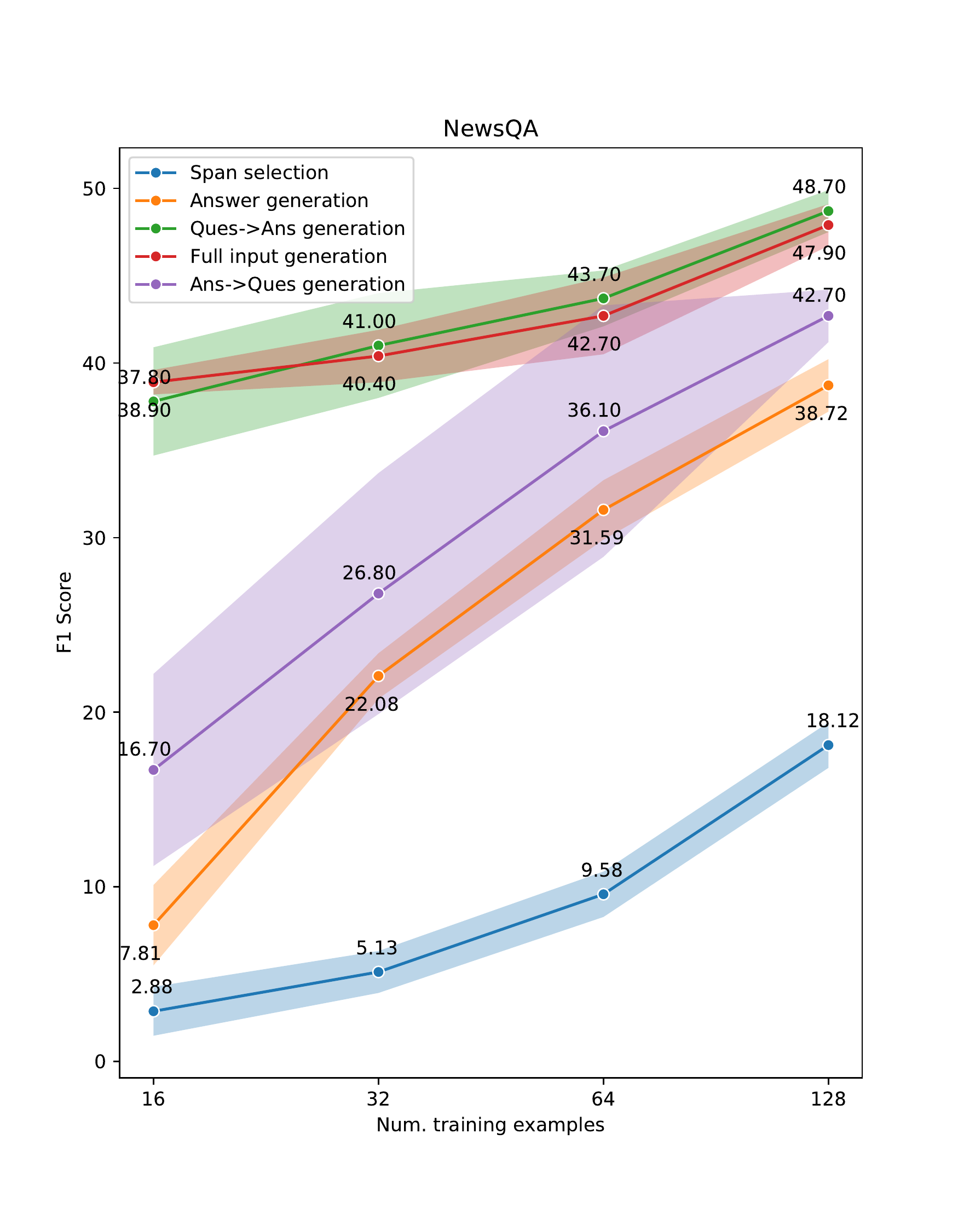}
         \label{fig:tydiqa}
     \end{subfigure}
        \caption{Comparison of fine-tuning objectives. The value on the markers indicates the mean. The beam width indicates the standard deviation.}
        \label{fig:ablations}
\end{figure*}


\subsection{Comparing the standard vs aligned fine-tuning framework}
First, we present results comparing the standard QA span-selection fine-tuning framework (BART) and our proposed fine-tuning framework (FewshotBART) that uses an input-output and an objective that is aligned with pre-training framework. We choose BART, two training set sizes (16, 128 examples) to illustrate this and present elaborate results across all configurations in a further section. Both BART and FewshotBART use the base version which contains 140M parameters. As seen in Table \ref{tab:alignment_results}, our proposed fine-tuning framework improves the F1 score significantly across all datasets in both the 16 example (an absolute gain of upto 48 F1 points and an average of 34.2 F1 points) and 128 example scenarios (an absolute gain of upto 39 F1 points and an average of 30.8 F1 points).

\subsection{Few-shot results}
Next, we present detailed experimental results (Table \ref{tab:fewshot_results}) obtained with our FewshotBART, FewshotBARTL, FewshotT5 models on several few-shot configurations across multiple datasets. For FewshotBARTL, the base model in FewshotBART is replaced with the larger 406M parameter model. We compare our models with RoBERTa, SpanBERT baselines and the recently proposed Splinter~\cite{DBLP:conf/acl/RamKBGL20}. The RoBERTa and SpanBERT are fine-tuned with the span-selection objective and we use the results for these models from~\cite{DBLP:conf/acl/RamKBGL20}.\\
We can see that FewshotBART, FewshotBARTL and FewshotT5 outperform the baselines by a big margin on almost all datasets. A few highlights are listed below:
\begin{itemize}
\item (a) Our best large model (FewshotBARTL) outperforms all other models by a big margin. Specifically, in a 16 example setting, it provides gains upto 61.2 F1 points in comparison to a similar-sized RoBERTa model that is fine-tuned with a span-selection objective.
\item (b) Our best comparable model to Splinter (in terms of model size) - FewshotBART outperforms it by upto 31.6 F1 points in a 16 example setting and upto 10.9 F1 points in a 128 example setting. TextbookQA dataset is one exception where Splinter is stronger.
\item (c) FewshotBART is stronger in a 16 example setting in comparison to FewshotT5. This difference starts fading in 32, 64 and 128 example settings. However, FewshotT5 still performs better on most of the datasets in comparison to Splinter in 16, 32 and 64 example settings.
\end{itemize}

\subsection{Choice of input-outputs and fine-tuning objectives}
In this section, we investigate the impact of changing the input-output design and the fine-tuning objective on the model performance (see Figure \ref{fig:ablations}). Given a question q, answer a and context c, we evaluate the following input-output choices and objectives:\\
\textbf{Span-selection}: This is the standard extractive question answering objective where the model is made to predict begin and end tokens corresponding to the answer in a given input I: \colorbox{gray!30}{Question: q [S] Context: c}\\
\textbf{Full input generation}: This is the objective where the model is made to predict the entire input including the masked answer span.\\ Input: \colorbox{gray!30}{Question: q Answer: <mask>. Context: c} \\
Target: \colorbox{gray!30}{Question: q Answer: a. Context: c}\\
\textbf{Question->Answer generation}: In this setting, the model is asked to generate only the question and the (masked) answer part of the input. The question tokens are followed by the answer tokens. The context tokens are not included in the fine-tuning objective.\\
Input: \colorbox{gray!30}{Question: q Answer: <mask>. Context: c} \\
Target: \colorbox{gray!30}{Question: q Answer: a.}\\
\textbf{Answer->Question generation}: This is similar to the \textbf{Question->Answer generation} objective with the difference being that the answer tokens are followed by the question tokens.\\
Input: \colorbox{gray!30}{Question: q Answer: <mask>. Context: c} \\
Target: \colorbox{gray!30}{Question: q Answer: a.}\\
\textbf{Answer generation}: This is the generation-based objective equivalent of the standard span-selection objective. The model is asked to generate the answer given an input question and context.\\ 
Input: \colorbox{gray!30}{Question: q Context: c} \\
Target T: \colorbox{gray!30}{a}\\

The results are plotted in Figure \ref{fig:ablations}. We choose SQuAD and NewsQA datasets for illustration. There are several key findings here, in the context of a few-shot QA setting (upto 128 examples):
\begin{itemize}
    \item (a) The fine-tuning objectives that are aligned with their pre-training objective (red, green, violet lines) show large gains over the standard span-selection fine-tuning objective (blue line). The answer generation objective (orange line) is superior to a span-selection based objective (blue line) on both the datasets.
    \item (b) The sequencing of question and answer tokens in the input-output has an impact on the performance with the specific sequencing of question followed by the answer being superior. 
    \item (c) The Question->Answer generation and Full input generation objectives show strong performance even when there are 16 examples. The gap between the span-selection and other objectives continue to be large even when there are 128 training examples.
\end{itemize}

\subsection{Multilingual results}

Here, we describe the results of applying FewshotmBART described in Section to a multilingual corpus TydiQA \cite{clark-etal-2020-tydi}. TydiQA consists of question answering datasets from 9 languages (Arabic, Bengali, English, Finnish, Indonesian, Korean, Russian, Swahili, Telugu). We compare this to the results from applying XLM-Roberta-large \cite{conneau-etal-2020-unsupervised} to the same dataset. We fine-tune XLM-Roberta-large using the standard span-selection objective used in extractive question answering tasks. The results are shown in Figure \ref{fig:tydiqa}.

FewshotmBART outperforms XLM-Roberta-large by an average of 32.96 absolute F1 points for training data sizes spanning from 2 to 64 examples.

\begin{figure}[H]
\centering
  \includegraphics[width=8cm,height=9cm]{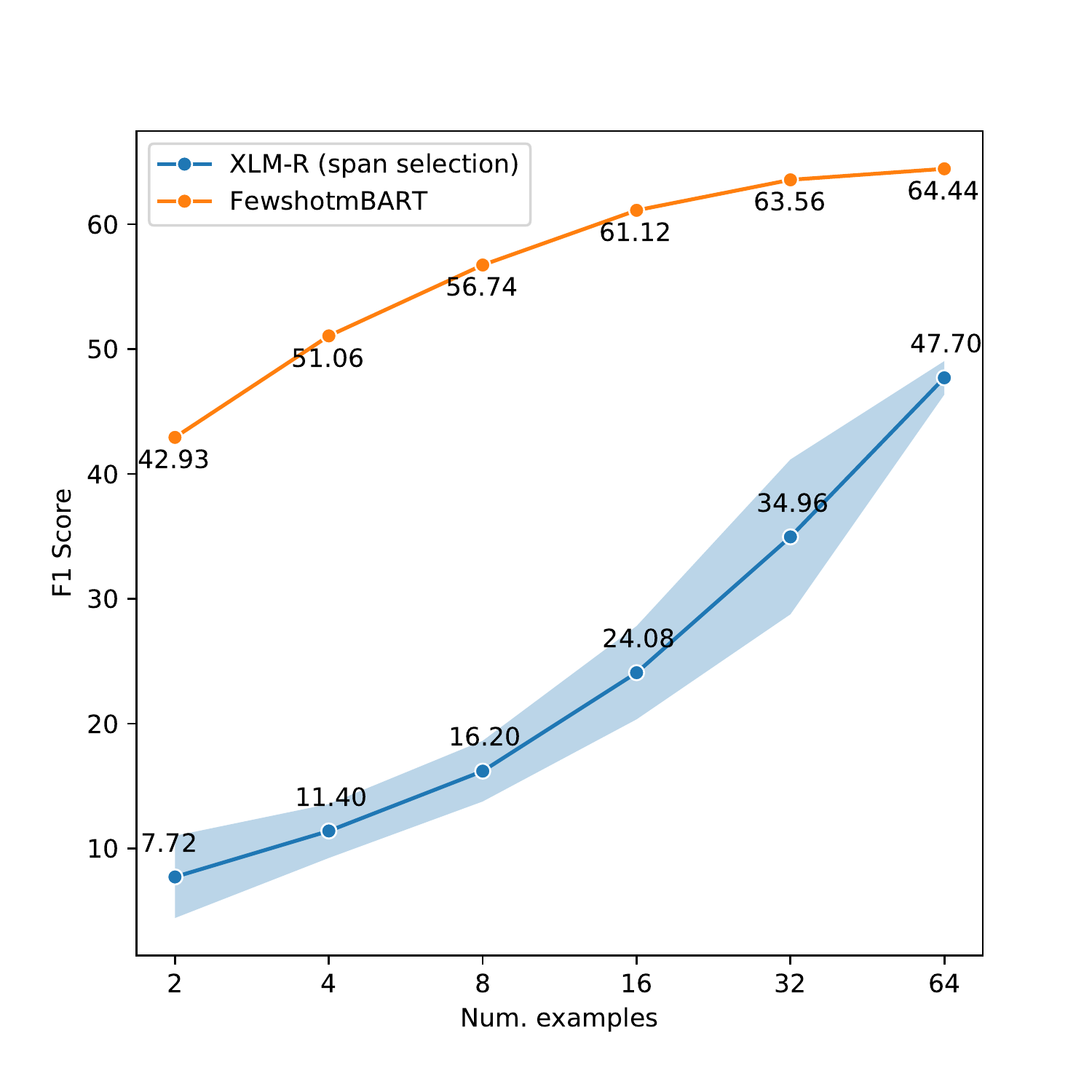}
  \caption{F1 comparison: XLM-Roberta vs FewshotmBART}
  \label{fig:tydiqa}
\end{figure}

\section{Related Work}
Question Answering (QA) is an active area of research in Natural Language Processing and the recent advances in pre-trained language models enabled lots of rapid progress in the field (\cite{devlin-etal-2019-bert}, \cite{NEURIPS2020_1457c0d6}, \cite{Bao2020UniLMv2PL}, \cite{JMLR:v21:20-074}). QA is also used as a format to cast several NLP problems~\cite{DBLP:journals/corr/abs-1806-08730}, ~\cite{chada-2019-gendered}. A common way to build a high performing question answering model is to fine-tune these pre-trained models on the entire training dataset - either via a span-extraction objective \cite{Lan2020ALBERT:}, \cite{Clark2020ELECTRA:}, \cite{Bao2020UniLMv2PL} or a span-generation objective \cite{JMLR:v21:20-074}. However, in this work, we explore a more challenging and a practical setting where only a handful of annotated training and development samples are available. Related to this, \cite{DBLP:journals/corr/abs-2108-05857} propose a decoding scheme for pre-trained language models that efficiently finds the most probable answer span in a passage.~\cite{DBLP:conf/acl/RamKBGL20} develop a new pretrained model that uses a recurring span selection objective suitable for QA tasks. They then fine-tune this customized pre-trained model on downstream QA tasks using the standard span selection objective. They argue that the existing strategy of fine-tuning large language models fail in a few-shot QA setting. In contrast, we take existing pre-trained text-to-text models BART \cite{lewis-etal-2020-bart}, T5 \cite{JMLR:v21:20-074} and simply modify their fine-tuning objective to build a stronger few-shot QA model. As our solution only relies on fine-tuning modifications, we are able to easily extend the framework to larger sized models and multilingual settings without having to build a new pre-trained model each time.\\
An alternative line of work that caters to building question answering models in low-data settings involves dataset synthesis~\cite{lewis-etal-2019-unsupervised}, ~\cite{alberti-etal-2019-synthetic}, ~\cite{puri-etal-2020-training}. ~\cite{lewis-etal-2019-unsupervised} generate pairs of synthetic context, question and answer triples by sampling context paragraphs from a large corpus of documents. Using these, they generate answer spans, mask the answer and use this cloze-style text to generate natural questions. To do this, they assume access to NLP tools such as named entity recognizer and part of speech tagger. \citet{puri-etal-2020-training} use a mix of BERT-based answer generation, GPT-2~\cite{Radford2019LanguageMA} based question generation and and roundtrip filtration to train an extractive QA model. They show promising results with larger scale models. However, the QA model is still fine-tuned on the entire dataset and this entire process including synthetic data generation is computationally expensive. Our work deviates from these by not relying on additional synthetic data, not assuming access to external NLP tools and using only a few training examples for fine-tuning.\\
There is also a connection of our work to some of the recent developments in few-shot learning for classification tasks that cast the problem as a mask-filling problem~\cite{schick-schutze-2021-exploiting}, ~\cite{gao-etal-2021-making}, ~\cite{Schick2020ItsNJ}. However, these solutions are geared towards classification tasks with a fixed set of classes. That assumption doesn't hold true for QA tasks.

\section{Conclusion}
We present an effective few-shot question answering (QA) system that combines the use of pre-trained text-to-text models and a fine-tuning framework aligned with their pre-training counterpart. Through experimental studies on various QA benchmarks and few-shot configurations, we show that this system can produce significant gains including in scenarios where the training data is extremely scarce (an absolute gain of 34 F1 points on average in comparison to the current standard of the fine-tuning framework). We also present extensions to multilingual and larger model settings and show that the gains translate well to these settings (eg:- up to an absolute 40 F1 point gain in comparison to XLM-Roberta + a span-selection objective). Through ablation studies, we study the impact of model size, fine-tuning objectives, input-output design and illustrate the factors leading to such strong gains. For future, as our framework doesn't explicitly enforce the answer to be a span in the input text, it'd be interesting to consider its applications to generative QA tasks.

\section*{Acknowledgements}
We thank Siddhant Garg and Karthik Gopalakrishnan for their valuable feedback and discussions. We thank Ori Ram for providing additional data on their Splinter experiments.

\bibliography{anthology,custom}
\bibliographystyle{acl_natbib}

\end{document}